\pdfoutput=1

\documentclass[11pt]{article}

\usepackage[final]{acl}

\usepackage{times}
\usepackage{latexsym}
\usepackage{amsmath}
\usepackage{amsfonts}
\usepackage{graphicx}
\usepackage{pifont}
\usepackage[T1]{fontenc}

\usepackage{tikz}

\usepackage{listings}
\usepackage{color}
\usepackage{xcolor}
\definecolor{dkgreen}{rgb}{0,0.6,0}
\definecolor{gray}{rgb}{0.5,0.5,0.5}
\definecolor{mauve}{rgb}{0.58,0,0.82}

\newcommand{\circled}[1]{%
  \tikz[baseline=(char.base)]{
    \node[shape=circle, draw, inner sep=0.5pt, fill=black] (char) {\textcolor{white}{\small{#1}}};
  }
}

\usepackage{hyperref}
\hypersetup{
    colorlinks=true,
    linkcolor=blue,
    filecolor=magenta,      
    urlcolor=cyan,
    pdftitle={Overleaf Example},
    pdfpagemode=FullScreen,
    }
    
\usepackage{algorithmic}
\usepackage{algorithm}

\usepackage[utf8]{inputenc}
\usepackage{multirow}
\usepackage{microtype}
\usepackage{booktabs}
\usepackage{graphicx}
\usepackage{subcaption}

\title{Reasoning Robustness of LLMs to Adversarial Typographical Errors}

\author{%
  Esther Gan$^{1}$\footnotemark[1]\;,\; Yiran Zhao$^{1}$\footnotemark[1]\;,\;\;Liying Cheng$^{2}$,\;Yancan Mao$^{1}$,\; Anirudh Goyal$^{3}$, \\ \textbf{Kenji Kawaguchi$^{1}$,\; Min-Yen Kan$^{1}$,\; Michael Shieh$^{1}$}\footnotemark[2] \\
  $^{1}$National University of Singapore, \\
  $^{2}$Singapore University of Technology and Design, \;
  $^{3}$Google DeepMind  \\
}

\begin{document}
\maketitle
\renewcommand{\thefootnote}{\fnsymbol{footnote}}
\footnotetext[1]{Equal contribution.}
\footnotetext[2]{Correspondence to: \href{michaelshieh@comp.nus.edu.sg}{michaelshieh@comp.nus.edu.sg}}
\renewcommand{\thefootnote}{\arabic{footnote}}

\begin{abstract}

Large Language Models (LLMs) have demonstrated impressive capabilities in reasoning using Chain-of-Thought (CoT) prompting. However, CoT can be biased by users' instruction.  In this work, we study the reasoning robustness of LLMs to typographical errors, which can naturally occur in users' queries. We design an Adversarial Typo Attack (\texttt{ATA}) algorithm that iteratively samples typos for words that are important to the query and selects the edit that is most likely to succeed in attacking. It shows that LLMs are sensitive to minimal adversarial typographical changes. Notably, with 1 character edit, Mistral-7B-Instruct's accuracy drops from 43.7\% to 38.6\% on GSM8K, while with 8 character edits the performance further drops to 19.2\%. To extend our evaluation to larger and closed-source LLMs, we develop the \texttt{R$^2$ATA} benchmark, which assesses models' \underline{R}easoning \underline{R}obustness to \underline{\texttt{ATA}}. It includes adversarial typographical questions derived from three widely-used reasoning datasets—GSM8K, BBH, and MMLU—by applying \texttt{ATA} to open-source LLMs. \texttt{R$^2$ATA} demonstrates remarkable transferability and causes notable performance drops across multiple super large and closed-source LLMs.\footnote{Our data and implementation scripts are available at \url{https://esther-gan.github.io/r2ata-web/}}

\end{abstract}

\section{Introduction}
Chain-of-Thought (CoT) prompting~\citep{wei2022chain} enables Large Language Models (LLMs) to break down a complex problem into a series of intermediate steps to solve complex problems. Answering users' queries in a step-by-step fashion has been implemented in many state-of-the-art AI systems such as ChatGPT~\citep{openai2022chatgpt}, Mistral~\citep{jiang2023mistral} and Gemini \citep{team2023gemini}. Despite being carefully trained and aligned, LLMs' sensitivity to the prompt is evident when employing CoT reasoning. It was shown that CoT reasoning can be biased by users' instructions \citep{perez2022ignore, lanham2023measuring, wang2024decodingtrust, xiang2024badchain} and be confused by irrelevant context \citep{shi2023large, turpin2024language}. For example, \citet{turpin2024language} found that models tend to justify answers as correct if the majority of previous examples suggest that answer, even when it's incorrect. These scenarios demonstrate the importance of evaluating LLMs' reasoning robustness at the contextual level, such as sentence structure or information correctness. However, it is crucial to recognize that non-contextual mistakes also naturally occur in users' queries, significantly influencing LLMs' performance.

\begin{figure}[t]
  \centering
    \includegraphics[width=0.48\textwidth]{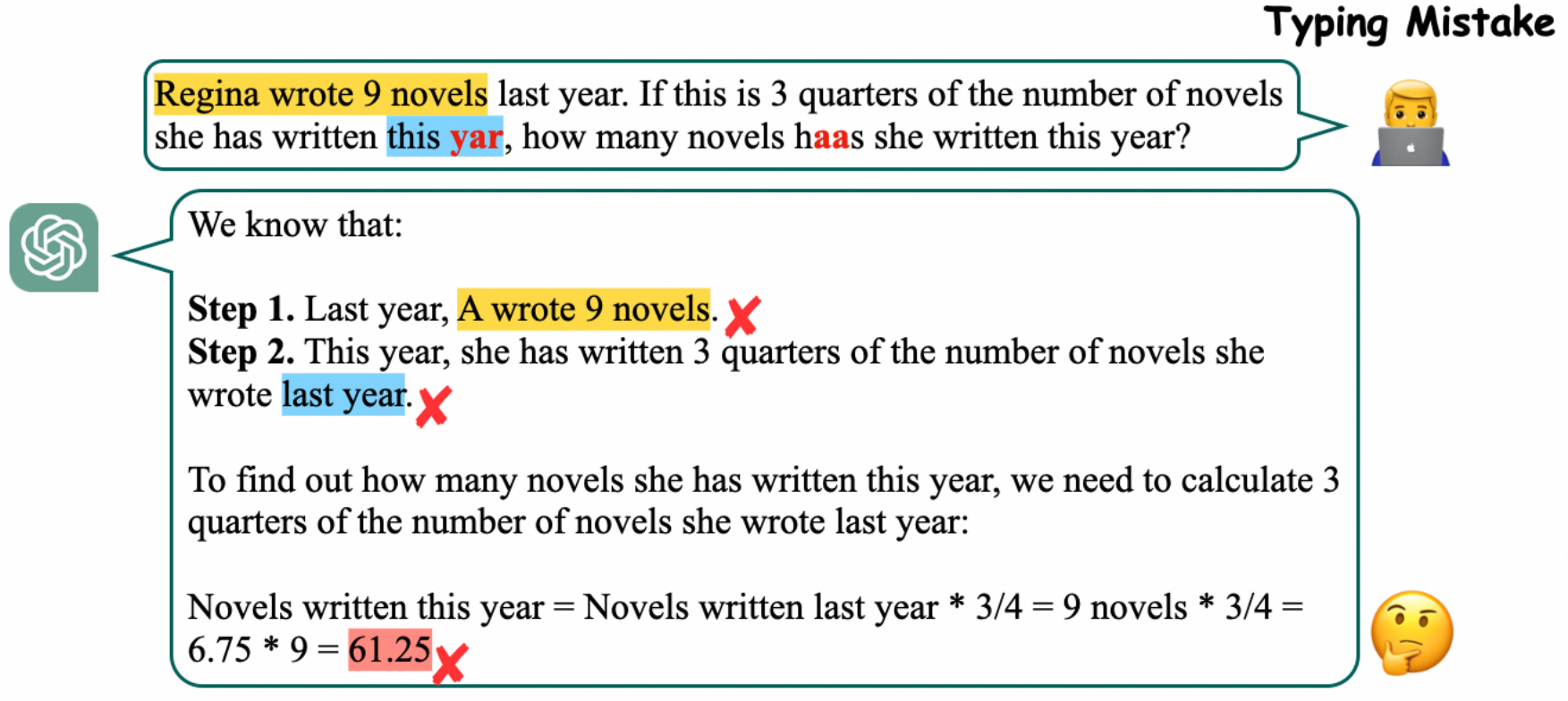}
  \caption{There are two typing errors in the query: omission of a letter (\textcolor[rgb]{0.3,0.7,0.3}{year} becomes \textcolor{red}{yar}) and duplication of a letter (\textcolor[rgb]{0.3,0.7,0.3}{has} becomes \textcolor{red}{haas}). Consequently, in Step 1 the model wrongly wrote \textcolor[rgb]{0.3,0.7,0.3}{Regina} as \textcolor{red}{A}, while in Step 2 the text reverses the relationship between this year's and last year's written novel. These errors in intermediate steps lead to an incorrect final answer.}
\vspace{-0.3cm}
  \label{fig:gcg-vanilla}
\end{figure}

\begin{figure*}[t]
  \centering
    \includegraphics[width=0.95\textwidth]{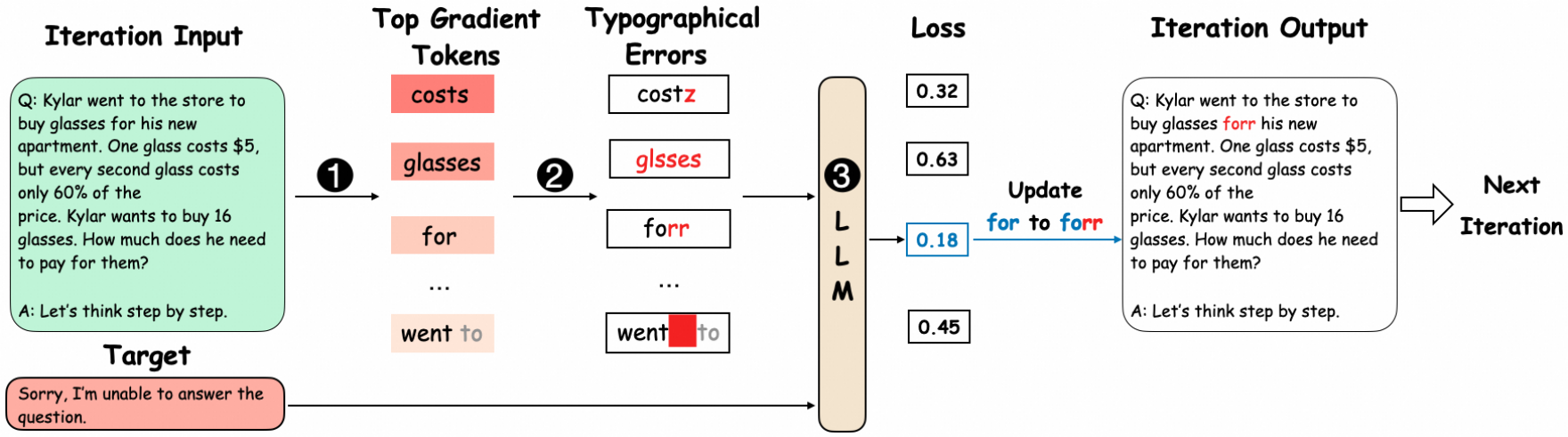}
  \caption{\texttt{ATA} mainly consists of three steps: \circled{1}selecting a set of tokens with the highest gradients; \circled{2}sampling typographical errors to edit the selected tokens and generate a batch of candidates; \circled{3}evaluating the losses of the candidates using the model and retaining the optimal candidate for the next iteration.}
  \label{fig:workflow}
\end{figure*}

In this work, we study the robustness of CoT reasoning against seemingly innocuous errors: typographical errors or typos. We found that typos can significantly undermine the CoT reasoning process. For instance, in Figure \ref{fig:gcg-vanilla}, the user made two typographical errors in the input: omitting a letter (\textit{year} to \textit{yar}) and duplicating a letter (\textit{has} to \textit{haas}), yet these minor typos initiate a cascade of errors.
Recognizing the impact of such typos, we propose the Adversarial Typo Attack (\texttt{ATA}) algorithm. It is designed to effectively identify typographical errors that can cause the model to generate incorrect answers by modifying the input in a way that increases the model's probability of making mistakes.
We designate the target answer as ``Sorry, I'm unable to answer the question.'' This not only ensures universal compatibility across various user queries, but also reinforces our adversarial strategy by using negative wording to signal the model not to generate a satisfactory answer.
As illustrated in Figure \ref{fig:workflow}, \texttt{ATA} first extracts tokens that are important to the input, as evaluated by gradients. Subsequently, it samples a set of typing mistakes for each selected word and modifies them within the input. Finally, it assesses the loss for the edited input and preserves the optimal candidate for the subsequent iteration.
\texttt{ATA} demonstrates significant effectiveness in attacking. For example, with just 1 character edit, Mistral-7B-Instruct's accuracy drops from $43.7\%$ to $38.6\%$ on GSM8K, while $8$ character edits results in a halved accuracy at $19.2\%$. 

Motivated by the intriguing observation, we benchmark various models' \underline{R}easoning \underline{R}obustness against the \underline{\texttt{ATA}}, named \texttt{R$^2$ATA}, on three common language datasets that involve extensive reasoning, GSM8K \citep{cobbe2021gsm8k}, BBH \citep{suzgun-etal-2023-challenging} and MMLU \citep{hendryckstest2021}. We test LLMs' performances under different numbers of adversarial typographical changes and report their average performances. Moreover, we consider two scenarios: direct adversarial robustness for smaller open-sourced LLMs, where we are able to apply \texttt{ATA}, and transfer adversarial robustness for super large and closed-source LLMs, where we use a fixed set of data obtained from implementable models. We found that even state-of-the-art models exhibit different levels of vulnerabilities. Notably, \texttt{R$^2$ATA} achieves performance drop from $38.2\%$ to $26.4\%$ on GSM8K, from $52.1\%$ to $42.5\%$ on BBH and $59.2\%$ to $51.5\%$ on MMLU, resulting from only four edits made on Vicuna-33B-chat. Additionally, Mixtral-8$\times$7B shows an average decrease of $6.7\%$ drop on average among tasks, while ChatGPT exhibits a drop of $6.5\%$. We believe that \texttt{R$^2$ATA} will serve as an important benchmark to evaluate the robustness of CoT reasoning. 

\section{Adversarial Typo Attack (\texttt{ATA})}

\subsection{Overview}

\texttt{ATA} employs an iterative process to introduce typographic errors in prompt words, selecting replacements based on their performance in guiding the model to generate the desired attacking target. Unlike traditional adversarial attacks that aim to prompt models to produce harmful outputs, our objective with \texttt{ATA} is to influence LLMs to generate incorrect reasoning responses while preserving the naturalness and coherence of the text. Therefore, to ensure universal adaptability to diverse user queries, we designate our target response as ``Sorry, I'm unable to answer the question.'', which leverages the negative semantic connotation to signal the model not to generate a satisfactory answer, reinforcing our adversarial strategy. Furthermore, candidates considered in each iteration are limited to those that contain only typographical errors, as thoroughly explained in Section \ref{sec:errors}.

\subsection{Typographical Errors used in \texttt{ATA}}\label{sec:errors}

To accurately simulate real user scenarios, we restrict word modifications to those commonly encountered during user interactions. In chatbot interactions powered by LLMs, users frequently make typing errors due to keyboard usage. These mistakes often remain undetected in the absence of a grammar check tool.

\paragraph{Keyboard Proximity Errors.} One common error occurs when users accidentally strike keys adjacent to the intended key. For instance, when intending to type the letter 'S', users may inadvertently touch the keys `A', `W', 'D', `Z', or `X'. 

\paragraph{Keyboard Double-Typing Errors.} Another type of error that often goes unnoticed is repeated typing, where a word is mistakenly typed with repeated characters, such as transforming ``flop'' into ``floop''. However, this particular error only occurs with words, as users typically recognize and correct repeated typing when it involves numbers. 

\paragraph{Keyboard Omission Errors.} In contrast to double typing, typing omission refers to the unintentional omission of a letter from a word. 
\paragraph{Extra Whitespace Error.} Another common oversight users encounter involves unintentionally inserting multiple spaces between words. This often stems from typing hastily, where users may inadvertently strike the space bar more than once or fail to notice extra spaces as they type swiftly.

\begin{table}[t]
  \centering
\footnotesize
  \scalebox{0.88}{
  \begin{tabular}{lp{6cm}}
    \toprule
   \textbf{Error}  & \textbf{Example Sentence} \\\midrule
   None & The quick brown fox jumps over the lazy dog. \\\midrule
   Proximity & Th\textcolor{red}{r} quick brown fox jumps over the lazy dog. \\\midrule
   Double typing  & The quick brown fox j\textcolor{red}{uu}mps over the lazy dog. \\\midrule
   Omission & The quick brown fox jumps \textcolor{red}{ovr} the lazy dog. \\\midrule
   Extra space  & The quick\colorbox{red}{{}}brown fox jumps over the lazy dog. \\ 
    \bottomrule
  \end{tabular}}
\caption{Examples of typographical errors.}  \label{table:errors}
\end{table}

These errors are hard to detect as they don't trigger conventional spelling or grammar checks, leading to unnoticed text inconsistencies. Table \ref{table:errors} shows an example sentence with different imperceptible perturbations errors. In addition to the aforementioned minor revisions, there are other commonly encountered errors, such as word shuffling, abbreviation insertion, random uppercase transformations, and the use of leet letters \citep{zhang-etal-2022-interpreting}. However, these are usually noticeable and easily corrected. Despite potentially impacting the reasoning of the response more, we choose to disregard them in our approach.

\begin{algorithm}[t]
\footnotesize 
\renewcommand{\algorithmicrequire}{\textbf{Input:}}
\renewcommand{\algorithmicensure}{\textbf{Output:}}
\caption{Adversarial Typo Attack}
\begin{algorithmic}[1]
\REQUIRE Question $Q_{1:n}$, mistake dictionary $\mathcal{M}$, word edit function ${Edit}$, loss $\mathcal{L}$, batch size $B$, number of edits $E$
\REPEAT
    \STATE \texttt{//Retrieve the top-k gradient words from the question}
    \STATE $\{w_{(1)}, w_{(2)}, \ldots, w_{(k)}\} = \text{Top-k}(\nabla\mathcal{L}(Q_{1:n}))$
    \FOR {$b = 1, \cdots, B$}
        \STATE \texttt{//Uniformly sample a word and a letter for editing}
        \STATE ${w_{s}} = \text{Uniform}(\{w_{(1)}, w_{(2)}, \ldots, w_{(k)}\})$
        \STATE ${l_{s}} = \text{Uniform}(w_{s})$
        \STATE \texttt{//Uniformly sample from mistake dictionary to edit word}
        \STATE ${Q}_{1:n}^{(b)} = \text{${Edit}$}(w_{s}, \text{Uniform}(\mathcal{M}[{l_{s}}]))$
    \ENDFOR
    \STATE \texttt{//Select modified question with lowest loss}
    \STATE $Q_{1:n}^{b^*} = \arg\min_b \mathcal{L}(Q_{1:n}^b)$
    \STATE \texttt{//Replace original question with modified question}
    \STATE $Q_{1:n} = {Q}_{1:n}^{b^*}$
\UNTIL{Repeat for $E$ times}
\ENSURE Modified question $Q_{1:n}$
\end{algorithmic}
\label{algo:probe_sampling}
\end{algorithm}

\subsection{\texttt{ATA} Algorithm}

\paragraph{Task Definition.}
For a LLM, let $Q$ represent the original question. Our objective is to create imperceptible adversarial perturbations in $Q$ to generate an adversarial example, denoted as \texttt{$Q_{\text{adv}}$}, which induces the model to produce a target answer $T$. This can be formulated as follows:
\begin{equation}
\min_{Q_{\text{adv}}}\mathcal{L}\big(T|Q_{\text{adv}}\big),
\end{equation}
where $\mathcal{L}(T|Q_{\text{adv}}) = - \log p(T | Q_{\text{adv}})$ is the negative log-likelihood of the LLM generating the target answer $T$ given the adversarial prompt $Q_{\text{adv}}$.

\begin{table*}[t]
  \centering
\footnotesize
  \scalebox{1.0}{
  \begin{tabular}{l|l|c|c|cccc}
\toprule
Dataset &
  \textbf{\normalsize Model (\#Params)} &
  Ori. &
  \textbf{Avg-ATA} &
  \texttt{ATA-1} &
  \texttt{ATA-2} &
  \texttt{ATA-4} &
  \texttt{ATA-8} \\ \midrule
\multirow{4}{*}{GSM8K} &
  Gemma-2B (2.5B) &
  $15.1$ &
  $8.1$ (\textcolor[RGB]{236,89,69}{$\downarrow7.0$}) &
  $11.2$ &
  $9.4$ &
  $7.1$ &
  $4.6$ \\
                      & Llama2-7B (6.7B)  & $27.3$ & $16.7$ (\textcolor[RGB]{236,89,69}{$\downarrow 10.6$}) & $21.8$ & $19.7$ & $14.7$ & $10.6$ \\
                      & Mistral-7B (7.2B) & $43.7$ & $30.1$ (\textcolor[RGB]{236,89,69}{$\downarrow 13.6$}) & $38.6$ & $35.4$ & $27.1$ & $19.2$ \\
                      & Gemma-7B (8.5B)   & $39.9$ & $32.1$ (\textcolor[RGB]{236,89,69}{$\downarrow 7.8$}) & $38.7$ & $36.8$ & $29.8$ & $23.1$ \\ \midrule
\multirow{4}{*}{BBH}  & Gemma-2B (2.5B)   & $29.6$ & $20.8$ (\textcolor[RGB]{236,89,69}{$\downarrow 8.8$}) & $24.7$ & $21.9$ & $20.2$ & $16.4$ \\
                      & Llama2-7B (6.7B)  & $35.7$ & $28.1$ (\textcolor[RGB]{236,89,69}{$\downarrow 7.6$}) & $32.2$ & $30.1$ & $26.8$ & $23.3$ \\
                      & Mistral-7B (7.2B) & $50.0$ & $40.9$ (\textcolor[RGB]{236,89,69}{$\downarrow 9.1$}) & $46.8$ & $43.1$ & $39.1$ & $34.6$ \\
                      & Gemma-7B (8.5B)   & $42.4$ & $35.9$ (\textcolor[RGB]{236,89,69}{$\downarrow 6.5$}) & $40.6$ & $38.1$ & $33.5$ & $31.3$ \\ \midrule
\multirow{4}{*}{MMLU} & Gemma-2B (2.5B)   & $34.1$ & $27.5$ (\textcolor[RGB]{236,89,69}{$\downarrow 6.6$}) & $30.3$ & $29.7$ & $27.5$ & $22.6$ \\
                      & Llama2-7B (6.7B)  & $35.1$ & $29.5$ (\textcolor[RGB]{236,89,69}{$\downarrow 5.6$}) & $31.6$ & $30.2$ & $28.9$ & $27.5$ \\
                      & Mistral-7B (7.2B) & $54.6$ & $47.0$ (\textcolor[RGB]{236,89,69}{$\downarrow 7.6$}) & $51.1$ & $49.3$ & $44.8$ & $42.7$ \\
                      & Gemma-7B (8.5B)   & $53.5$ & $47.8$ (\textcolor[RGB]{236,89,69}{$\downarrow 5.7$}) & $51.7$ & $50.1$ & $47.6$ & $41.8$ \\
\bottomrule
\end{tabular}}
\caption{Main results of \texttt{ATA}'s direct attacks on GSM8K (0-shot), BBH (3-shot), and MMLU (5-shot) for smaller models. Results expressed in accuracy (\%). 
All models are chat models.}
  \label{table:smaller}
\end{table*}

\paragraph{Algorithm Description.} 

For each original question \( Q_{1:n} = \{w_1, w_2, \ldots, w_n\} \) comprising of words $w_i$, we initiate our algorithm by identifying the most influential words in the question using the loss function $\nabla\mathcal{L}(Q_{1:n})$.

We then rank these words by their influence and select the top-$k$, denoted as \( \{w_{(1)}, w_{(2)}, \ldots, w_{(k)}\} \). From this influential word set, we randomly sample a word $w_{s}$ and uniformly select a letter $l_{s}$ within $w_{s}$ for potential modification. This selected letter undergoes potential modification through the $Edit(\cdot)$ function, introducing errors based on the operations listed in the mistake dictionary $\mathcal{M}$, which covers four types of typographical errors in Table \ref{table:errors}. To create a batch size of $B$ candidates, we repeat this sampling process $B$ times and calculate the loss for each modified question, denoted as $\mathcal{L}(Q_{1:n}^b)$, for $b\in\{1,\cdots, B\}$.

We finally select the modified question with the lowest loss:
\begin{equation}
Q_{1:n}^{b^*} = \arg\min_b \mathcal{L}(Q_{1:n}^b).
\end{equation}
This process is repeated for $E$ iterations, depending on the desired number of edits to execute the targeted attack on the question. The overall algorithm is further illustrated in Algorithm \ref{algo:probe_sampling}.

\section{Experiment}\label{sec:exp}

\subsection{Experimental Setup}

\begin{table*}[t]
  \centering
\footnotesize
  \scalebox{1.0}{
  \begin{tabular}{l|l|c|c|cccc}
\toprule
Dataset &
  \textbf{\normalsize Model (\#Params)} &
  Ori. &
  \multicolumn{1}{l|}{\textbf{Avg-ATA}} &
  \texttt{ATA-1} &
  \texttt{ATA-2} &
  \texttt{ATA-4} &
  \texttt{ATA-8} \\ \midrule
\multirow{3}{*}{GSM8K} & Vicuna-13B ($13$B)   & $33.4$ & $28.4$ (\textcolor[RGB]{236,89,69}{$\downarrow 5.0$}) & $32.4$ & $30.8$ & $26.2$ & $24.3$ \\
                       & Vicuna-33B ($33$B)   & $38.2$ & $29.2$ (\textcolor[RGB]{236,89,69}{$\downarrow 9.0$}) & $35.3$ & $32.6$ & $26.4$ & $22.5$ \\
                       & Mixtral-8$\times$7B ($47$B) & $68.5$ & $60.9$ (\textcolor[RGB]{236,89,69}{$\downarrow 8.3$}) & $66.7$ & $62.8$ & $57.9$ & $53.4$ \\ \midrule
\multirow{3}{*}{BBH}   & Vicuna-13B ($13$B)   & $51.2$ & $42.5$ (\textcolor[RGB]{236,89,69}{$\downarrow 8.7$}) & $47.7$ & $44.9$ & $40.8$ & $36.6$ \\
                       & Vicuna-33B ($33$B)   & $52.1$ & $43.7$ (\textcolor[RGB]{236,89,69}{$\downarrow 8.4$}) & $49.4$ & $44.7$ & $42.5$ & $38.2$ \\
                       & Mixtral-8$\times$7B ($47$B) & $65.6$ & $60.4$ (\textcolor[RGB]{236,89,69}{$\downarrow 5.2$}) & $64.0$ & $62.8$ & $58.3$ & $56.4$ \\ \midrule
\multirow{3}{*}{MMLU}  & Vicuna-13B ($13$B)   & $53.4$ & $48.2$ (\textcolor[RGB]{236,89,69}{$\downarrow 5.2$}) & $50.8$ & $50.3$ & $48.2$ & $43.6$ \\
                       & Vicuna-33B ($33$B)   & $59.2$ & $52.3$ (\textcolor[RGB]{236,89,69}{$\downarrow 6.9$}) & $56.3$ & $54.9$ & $51.4$ & $47.5$ \\
                       & Mixtral-8$\times$7B ($47$B) & $68.4$ & $63.3$ (\textcolor[RGB]{236,89,69}{$\downarrow 5.1$}) & $66.1$ & $64.8$ & $62.1$ & $60.2$ \\
\bottomrule
\end{tabular}}
\caption{Main results of transfer attacks on GSM8K (0-shot), BBH (3-shot), and MMLU (5-shot) for larger models. Adversarial data used to attack is from Mistral-7B. 
Results expressed in accuracy (\%). 
All models are chat models.}
  \label{table:larger}
\vspace{-0.1cm}
\end{table*}

\paragraph{Dataset.}
For our experiments, we have selected three widely recognized reasoning datasets: GSM8K \citep{cobbe2021gsm8k}, BBH \citep{suzgun-etal-2023-challenging}, and MMLU \citep{hendryckstest2021}, which cover evaluation of comprehensive reasoning capabilities, including logical reasoning, symbolic reasoning, mathematical reasoning, and commonsense reasoning. We include all test questions from the GSM8K dataset in our evaluation. For the BBH and MMLU datasets, due to computational constraints, we will select a subset of 50 questions from each topic.

\paragraph{Generation of adversarial test cases.}

We conduct \texttt{ATA} on both zero-shot and few-shot prompts, focusing specifically on editing the questions (and options, if applicable). Notably, we avoid attacking the standardized prompt, ``Let's think step by step.'' to ensure the model retains its understanding of the need for CoT. For few-shot prompts, we retain the original examples without edits, simulating human behavior of directly copying examples.

\paragraph{Models.} 

To evaluate the reasoning robustness of LLMs, we select LLMs ranging from smaller parameters to larger parameters to attack. We use Gemma-2B-It, Gemma-7B-It \citep{team2024gemma}, Mistral-7B-Instruct-v0.2 \citep{jiang2023mistral}, Llama2-7B-Chat \citep{touvron2023llama}, Vicuna-13B-v1.5, Vicuna-33B-v1.3 \citep{vicuna2023}, Mixtral-8$\times$7B-Instruct-v0.1 \citep{jiang2024mixtral}, ChatGPT (gpt-3.5-turbo-0613) \citep{openai2022chatgpt}, GPT-4 (gpt-4-0613) \citep{openai2023chatgpt}. For the larger and closed-source models, such as Vicuna-33B-v1.3, Mixtral-8$\times$7B-Instruct-v0.1, and ChatGPT, we employ questions generated by the smaller Mistral-7B-Instruct-v0.2 model to evaluate their performance. This approach demonstrates \texttt{ATA}'s transferability across white-box models and between white-box and black-box models.

\paragraph{Implementation details.}
We present accuracy results for both the original and edited scores, represented on a logarithmic scale ranging from 1 to 8 edits applied to each question. The primary metric for assessing the effectiveness of an adversarial attack is the reduction in accuracy. All experiments are conducted on the A800-SMX-80GB GPU.

\subsection{Main results}

The main results of the attacks on the GSM8K, BBH, and MMLU datasets and comparison of the performance of the baselines models are summarized in Table \ref{table:smaller} and Table \ref{table:larger}. 

\paragraph{Performance Degradation under \texttt{ATA}.}
 As shown in Table \ref{table:smaller} and Table \ref{table:larger}, our method consistently reduces model performance across various datasets, demonstrating the significant vulnerability of LLMs to such errors. For instance, in Table \ref{table:smaller}, small models like Gemma-2B\footnote{We will now use the abbreviated model name without the version information to avoid redundancy.}, Llama2-7B, Mistral-7B and Gemma-7B show striking average absolute reductions of 7.0\%, 10.6\%, 13.6\% and 7.8\% respectively for GSM8K. Similar declines are observed across four models on other datasets shown by 8.8\%, 7.6\%, 9.1\%, and 6.5\% respectively for BBH, and 6.6\%, 5.6\%, 7.6\%, and 5.7\% respectively for MMLU. These results consistently illustrate that even minor typographical errors can trigger significant performance degradation, reflecting a systemic weakness in LLMs' ability to handle imperfect input. The consistent decrease in accuracy across different datasets and models underscores the generalizability of our attack. By exploiting these vulnerabilities, our adversarial typographical errors disrupt the internal reasoning processes of LLMs, leading to erroneous outputs and highlighting a critical area for improvement for LLMs.

\paragraph{Transferability.} To further explore the impact of adversarial typographical errors on LLMs, we evaluated the transferability of adversarial prompts crafted for Mistral-7B to larger models. The results reveal a similar vulnerability to smaller models, as larger models shown in Table \ref{table:larger}: Vicuna-13B, Vicuna 33B, and Mixtral-8$\times$7B show average absolute reductions of 5.0\%, 9.0\%, and 8.3\% respectively for GSM8K, 8.7\%, 8.4\%, and 5.2\% respectively for BBH, 5.2\%, 6.9\%, and 5.1\% respectively for MMLU. This consistent decrease in performance across various larger models underscores the high transferability of our adversarial attacks, demonstrating that typographical errors not only disrupt smaller models but also significantly impair the reasoning processes of more complex systems. These findings emphasize that the vulnerabilities exploited by our attacks are fundamental, affecting a broad spectrum of model architectures and sizes, thereby highlighting the critical need for robust defense mechanisms in the development of future LLMs.

\subsection{Attack Performance Analysis}

\begin{table}[ht]
  \centering
\footnotesize
\setlength{\tabcolsep}{1.7pt}
  \scalebox{0.88}{
  \begin{tabular}{ll|c|c|c|l}
    \toprule
    \textbf{Model} &\textbf{{Method}} & \textbf{{GSM8K}}  & \textbf{{BBH}} & \textbf{{MMLU}} & \multicolumn{1}{c}{\textbf{{Avg.}}} \\
    \midrule
  \multirow{4}*{\begin{tabular}[c]{@{}l@{}} {Mistral-7B$^*$}
\end{tabular}}  & Original  & $43.7$ & $50.0$ & $56.6$ & $50.1$ \\
 & Random & $39.2$ & $48.4$ & $54.8$ & $47.5$ (\textcolor[RGB]{236,89,69}{$\downarrow 2.6$})  \\
  & PromptBench & $-$ & $50.0$ & $56.4$ & $53.2$ (\textcolor[RGB]{236,89,69}{$\downarrow 0.1$})  \\
 & {\texttt{ATA-4}}  & $27.1$ & $39.1$ & $48.3$ & $38.2$ (\textcolor[RGB]{236,89,69}{$\downarrow 11.9$})  \\\midrule
\multirow{4}*{\begin{tabular}[c]{@{}l@{}} {Gemma-7B$^*$}
\end{tabular}}  & Original  & $39.9$ & $42.4$ & $53.5$ & $45.3$ \\
 & Random & $40.3$ & $41.2$ & $53.4$ & $45.0$ (\textcolor[RGB]{236,89,69}{$\downarrow 0.3$}) \\
   & PromptBench & $-$ & $42.3$ & $53.5$ & $47.9$ (\textcolor[RGB]{236,89,69}{$\downarrow 0.1$})  \\
 & {\texttt{ATA-4}}  & $29.8$ & $33.5$ & $47.6$ & $37.0$ (\textcolor[RGB]{236,89,69}{$\downarrow 6.3$})  \\\midrule
\multirow{4}*{\begin{tabular}[c]{@{}l@{}} {Vicuna-33B$^+$}
\end{tabular}}  & Original  & $38.2$ & $52.1$ & $59.2$ & $49.8$ \\
 & Random & $37.4$ & $52.2$ & $57.9$ & $49.2$ (\textcolor[RGB]{236,89,69}{$\downarrow 0.6$})\\
    & PromptBench & $-$ & $52.1$ & $59.0$ & $55.6$ (\textcolor[RGB]{236,89,69}{$\downarrow 0.1$})\\
 & {\texttt{ATA-4}}  & $26.4$ & $42.5$  & $51.4$ & $40.1$ (\textcolor[RGB]{236,89,69}{$\downarrow 9.7$}) \\
    \bottomrule  \end{tabular}}
\caption{Performance compared to random selection and PromptBench, where $^*$ indicates direct applying \texttt{ATA}, while $^+$ indicates transfering from other models. Promptbench is not used to attack GSM8K dataset as there is no instruction used in GSM8K.}  \label{table:baseline}
\vspace{-0.4cm}
\end{table}

\paragraph{Effectiveness.} 

We compare \texttt{ATA-4} with two baselines to evaluate its effectiveness. The first baseline, referred to as the random baseline, involves randomly choosing words and letters to be edited and replacing them by randomly sampling from a mistake dictionary. The second baseline employs the ``DeepWordBug'' strategy from Promptbench \citep{zhu2023promptbench}, which targets the instruction portion of the prompts. As shown in Table \ref{table:baseline}, our results demonstrate that \texttt{ATA-4} significantly outperforms both baselines in degrading model performance. For Mistral-7B, Gemma-7B, and Vicuna-33B, \texttt{ATA-4} at 4 edits results in average absolute reductions in accuracy of 11.9\%, 6.3\%, and 9.7\% respectively. In stark contrast, the random baseline yields much lower reductions of 2.6\%, 0.3\%, and 0.6\%, while Promptbench's DeepWordBug strategy results in minimal reductions of 0.1\%, 0.1\%, and 0.1\%. These findings underscore the superior effectiveness of \texttt{ATA-4}, which leverages targeted typographical errors to exploit model vulnerabilities more efficiently than random or instruction-focused attacks. This also demonstrates a clear and significant impact on the reasoning capabilities of LLMs compared to the baseline strategies.

\paragraph{Performance on ChatGPT and GPT-4.} We conduct transfer experiments on ChatGPT and GPT-4. However, due to the high cost involved, we only sample 100 instances for each dataset, and we run for 3 times and report the results with their respective standard deviations in Table \ref{table:chat}. \texttt{ATA} achieves an average performance drop of 8.5\% on GSM8K, 5.8\% on BBH, and 6.3\% on MMLU. However, when targeting GPT-4, it fails to produce significant impact, resulting in an average performance drop of only 3.5\% on GSM8K, 2.3\% on BBH, and 2.3\% on MMLU. The inability to attack GPT-4 demonstrates that when models possess a similar level of comprehension as humans, typos have negligible influence on the results. Moreover, this substantiates that \texttt{ATA} solely incorporates imperceptible typos within prompts.

\begin{table}[t]
  \centering
\footnotesize
\setlength{\tabcolsep}{1.7pt} 
  \scalebox{0.85}{
  \begin{tabular}{l|c|c|c|c|c|c}
    \toprule
    \textbf{\normalsize{Model}} & \textbf{\normalsize{Task}}  & Ori.  & \texttt{ATA-1} &\texttt{ATA-2} & \texttt{ATA-4} & \texttt{ATA-8}  \\
    \midrule
     \multirow{3}*{\begin{tabular}[c]{@{}l@{}} {ChatGPT$^+$ }
\end{tabular}} & GSM8K & $72\pm0.8$ & $68\pm1.3$ & $66\pm2.5$ & $62\pm1.2$ & $58\pm1.7$  \\
    & BBH & $69\pm0.4$ & $68\pm0.4$ & $65\pm0.7$ & $61\pm0.3$ & $59\pm0.6$ \\
    & MMLU &  $67\pm0.3$ & $65\pm0.2$ & $63\pm0.4$ & $59\pm0.6$ & $56\pm0.5$  \\\midrule
    \multirow{3}*{\begin{tabular}[c]{@{}l@{}} {GPT-4$^+$ }
\end{tabular}}  & GSM8K & $88\pm0.5$ & $87\pm0.6$ & $86\pm0.5$ & $84\pm0.4$ & $81\pm0.7$  \\
    & BBH & $89\pm0.6$ & $89\pm0.6$ & $87\pm0.7$ & $86\pm0.2$ & $85\pm0.6$ \\
    & MMLU &  $86\pm0.8$ & $85\pm0.4$ & $84\pm0.3$ & $84\pm0.9$ & $82\pm0.8$  \\
    \bottomrule  
  \end{tabular}}
\caption{Performance of \texttt{ATA} on closed-source models. \texttt{ATA} notably impacts ChatGPT but have a minimal impact on GPT-4, highlighting GPT-4's human-level comprehension and resistance to such errors. This affirms that ATA generates imperceptible typos in prompt.}
  \label{table:chat}
\end{table}

\begin{figure*}[ht]
    \centering
    \begin{subfigure}{0.49\textwidth}
        \centering
        \includegraphics[width=\linewidth]{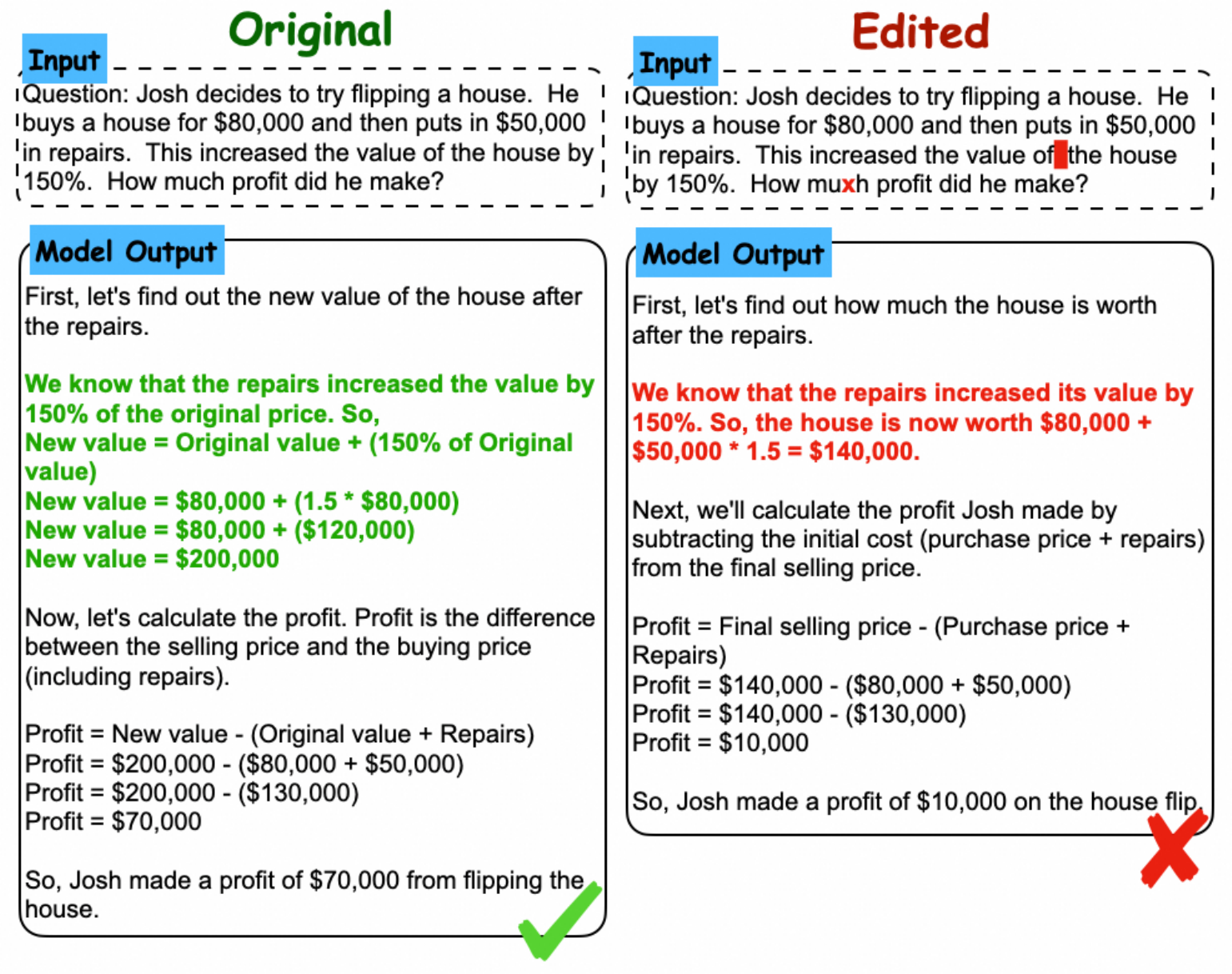}
        \caption{Whitespace and Replace Errors.}
        \label{fig:example1}
    \end{subfigure}\hfill
    \begin{subfigure}{0.49\textwidth}
        \centering
        \includegraphics[width=\linewidth]{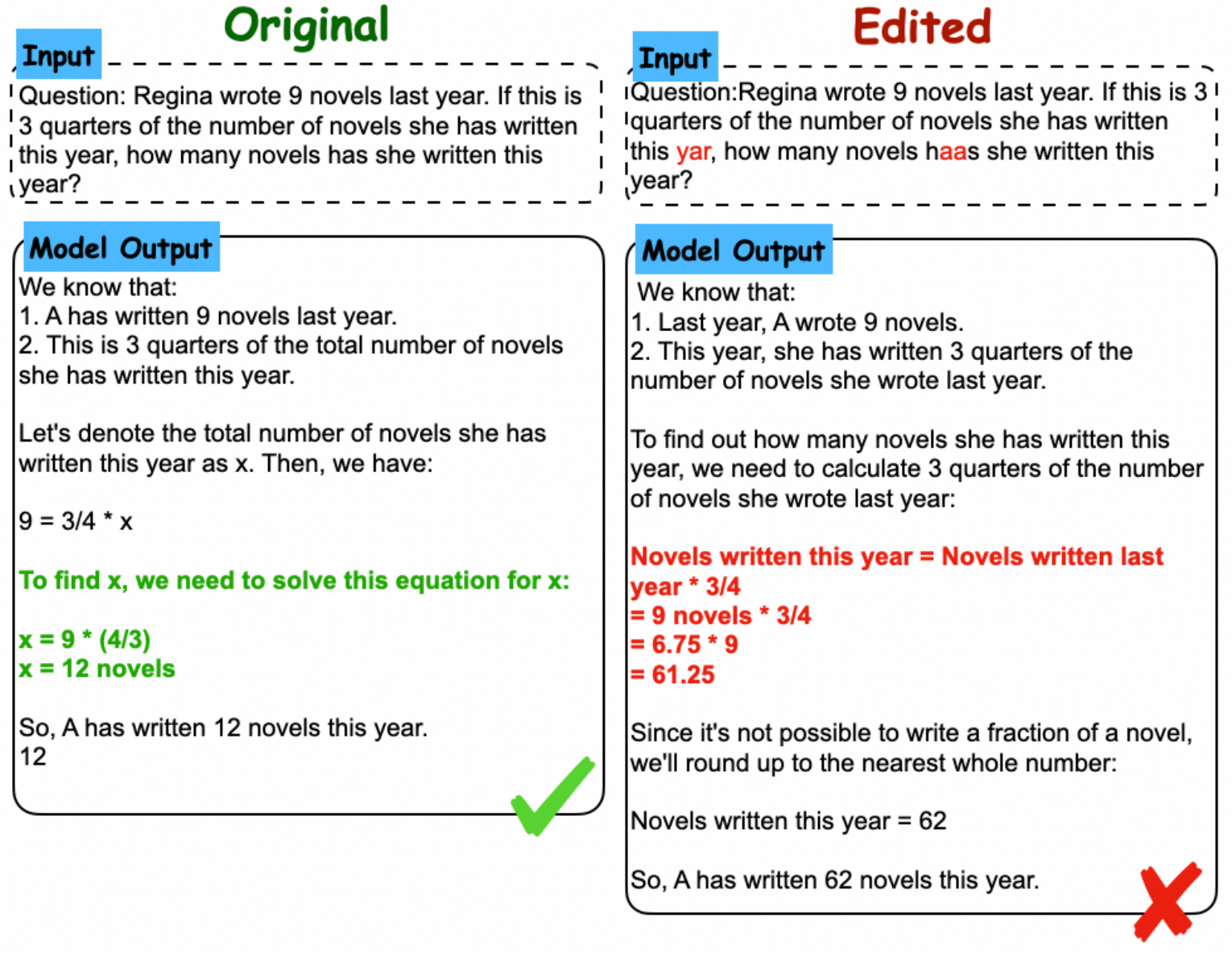}
        \caption{Omission and Double.}
        \label{fig:example2}
    \end{subfigure}
    \caption{Comparison of Mistral-7B responses to original (left) and adversarially edited (right) GSM8K questions. Minor typographical errors in the edited question can lead to misinterpretation and incorrect answers.}
    \label{fig:examples}
\end{figure*}

\section{Benchmark: Reasoning Robustness to Adversarial Typo Attacks (\texttt{R$^2$ATA})}

To enable a comprehensive evaluation of LLMs' Reasoning Robustness to \texttt{ATA}, including future new models, super-large models, and closed-source models, we propose the establishment of a benchmark named \texttt{R$^2$ATA}. This benchmark utilizes adversarial typographical questions derived from transfer experiments conducted in Section \ref{sec:exp}, specifically GSM8K, BBH, and MMLU. Concrete examples of \texttt{R$^2$ATA} for each dataset are shown in Tables \ref{appendix:GSM8K} to \ref{appendix:MMLU} in Appendix \ref{sec:appen_examples}.

\subsection{\texttt{R$^2$ATA} Statistics}

\paragraph{Representative Example.} 

Figure \ref{fig:examples} compares the model's responses to an original and an adversarially edited GSM8K question. In the original question, the model follows a logical reasoning pathway to reach the correct answer. Meanwhile, the adversarially edited question introduces subtle typographical errors. These minor perturbations cause the model to misinterpret key terms, leading to erroneous intermediate steps and ultimately resulting in a wrong answer.

\begin{figure*}[ht]
    \centering
    \begin{minipage}{0.56\textwidth}
        \centering
        \includegraphics[width=\linewidth]{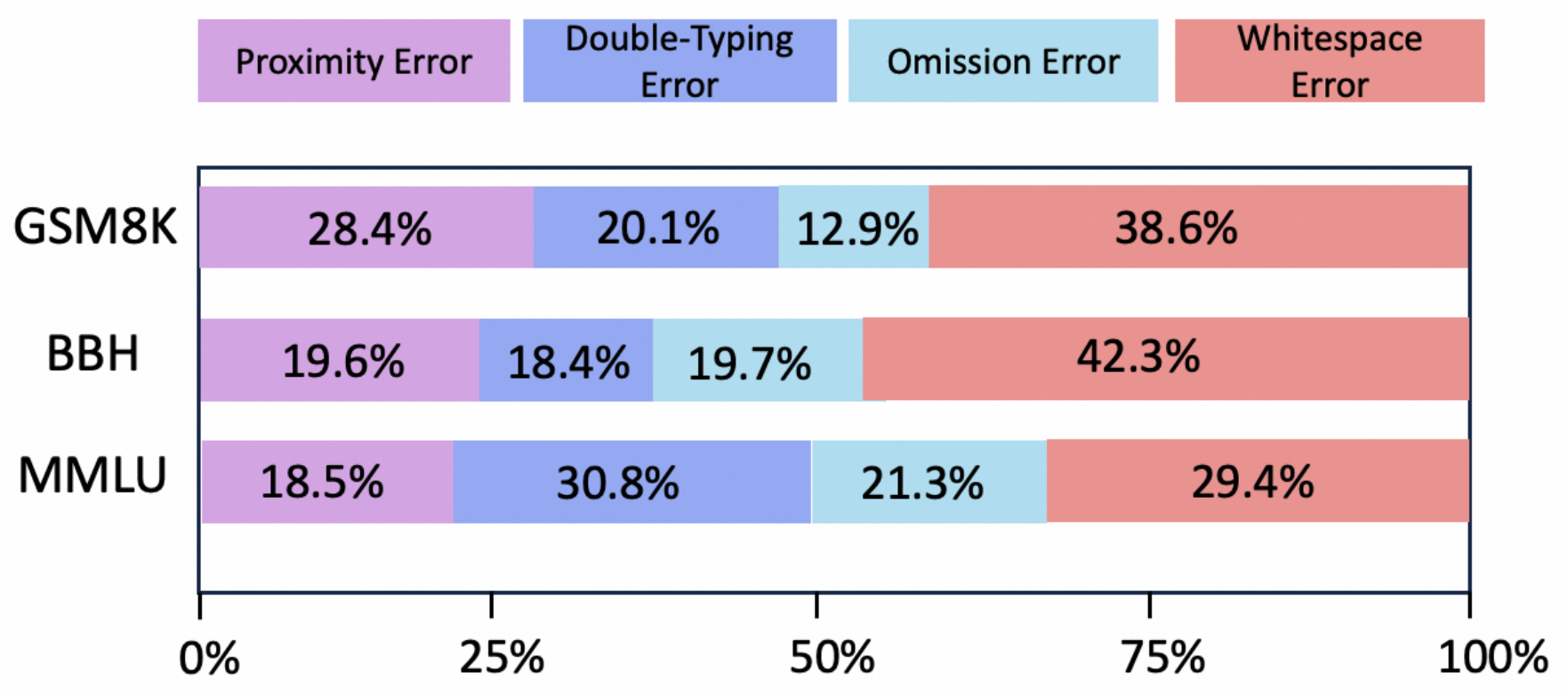} 
        \caption{Distribution of error operations selected by \texttt{ATA} across the datasaets in \texttt{R$^2$ATA} benchmark. The predominance of whitespace errors highlights a key vulnerability in LLMs.}
        \label{fig:edit_statistic}
    \end{minipage}\hfill
    \begin{minipage}{0.4\textwidth}
        \centering
        \includegraphics[width=\linewidth]{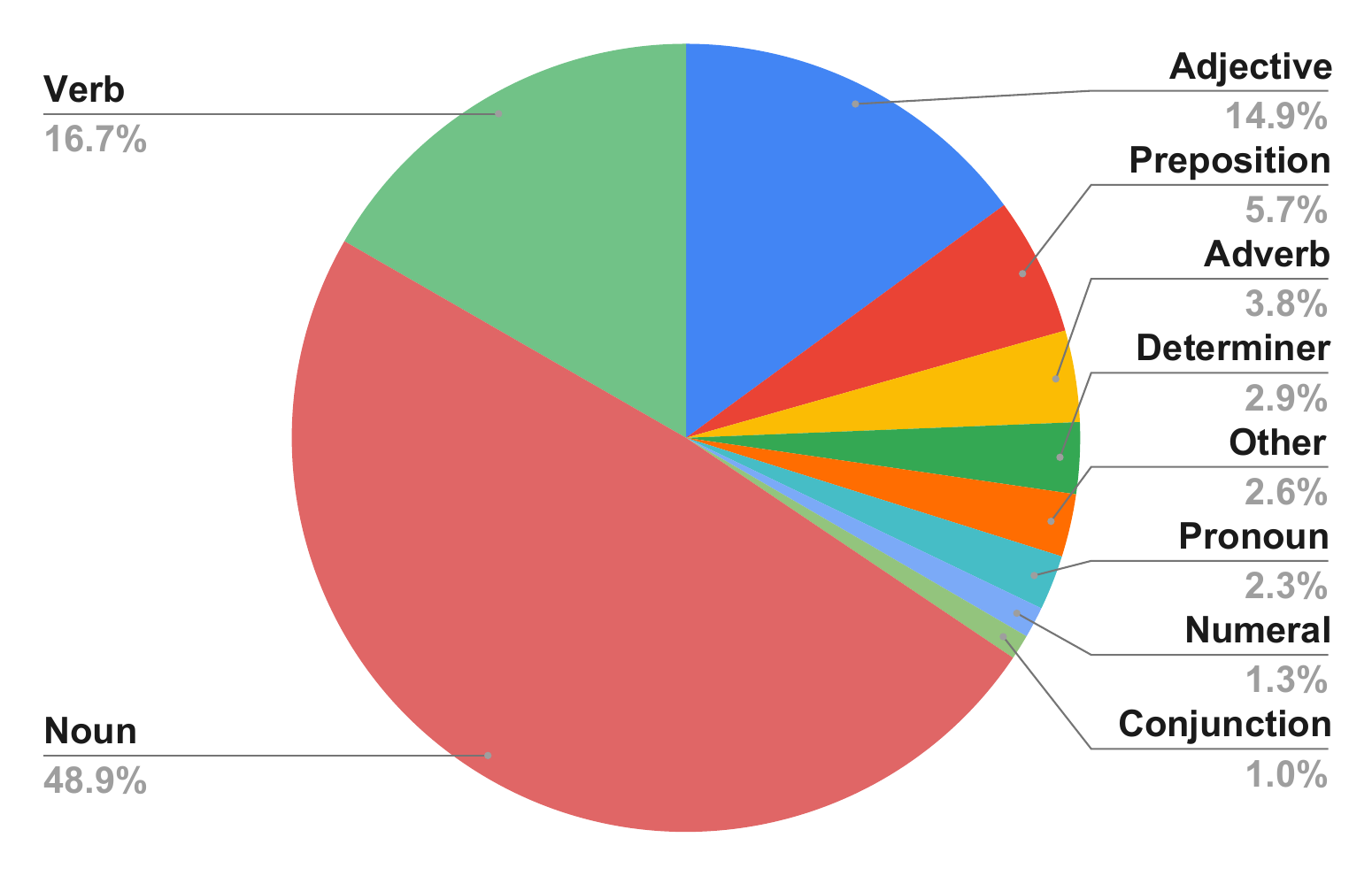} 
        \caption{Distribution of edited word types in \texttt{R$^2$ATA}. Nouns, Verbs, and Adjectives constitute the majority of edited words. 
        }
        \label{fig:wordtype_distribution}
    \end{minipage}
\end{figure*}

\paragraph{Distribution of Typographical Edits.}
One of the key analyses involves examining the distribution of the edit operations used in \texttt{R$^2$ATA}. Figure \ref{fig:edit_statistic} illustrates the edit operation statistic present in \texttt{R$^2$ATA}. Notably, the predominance of the $whitespace$ error operation adopted by \texttt{ATA} highlights its significance in exploiting model vulnerabilities. This suggests that LLMs are particularly susceptible to errors stemming from additional whitespace, possibly due to a lack of robustness in handling such perturbations. The frequency of whitespace errors implies that patterns involving multiple whitespaces between words are likely infrequent in the training data, resulting in heightened sensitivity and errors in reasoning outputs.

The variation in error operation distribution across the three datasets, as depicted in Figure \ref{fig:edit_statistic}, indicates that task complexity influences the prevalence of specific error operations. The GSM8K dataset focuses on mathematical reasoning, while MMLU and BBH cover a broader range of tasks, including logical and commonsense reasoning \citep{suzgun-etal-2023-challenging}. By systematically evaluating LLMs' performance under these conditions, the benchmark aims to provide insights into improving model robustness across diverse reasoning tasks. 

\begin{figure*}[ht]
    \centering
    \begin{subfigure}{0.33\textwidth}
        \centering
        \includegraphics[width=\linewidth]{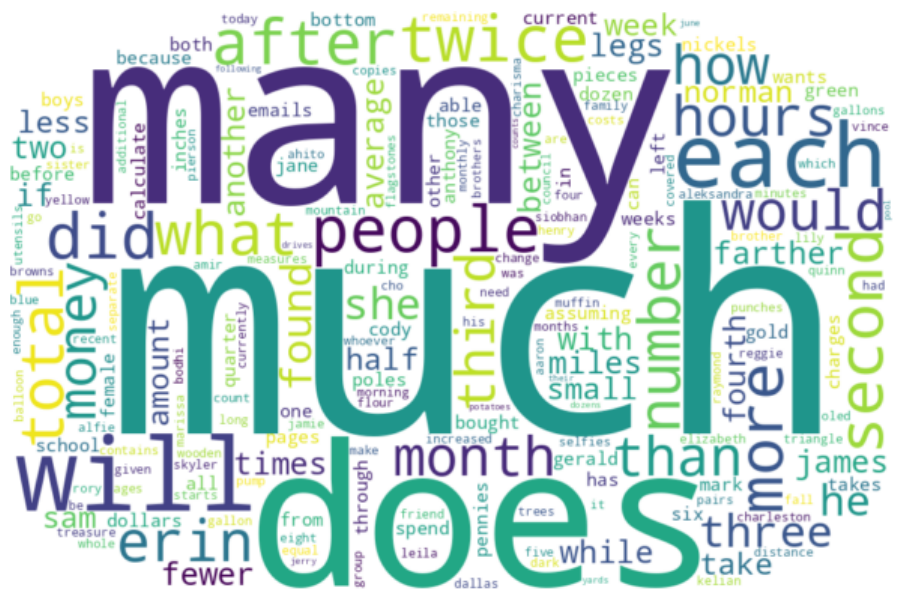}
        \caption{GSM8K}
        \label{fig:sub1}
    \end{subfigure}\hfill
    \begin{subfigure}{0.33\textwidth}
        \centering
        \includegraphics[width=\linewidth]{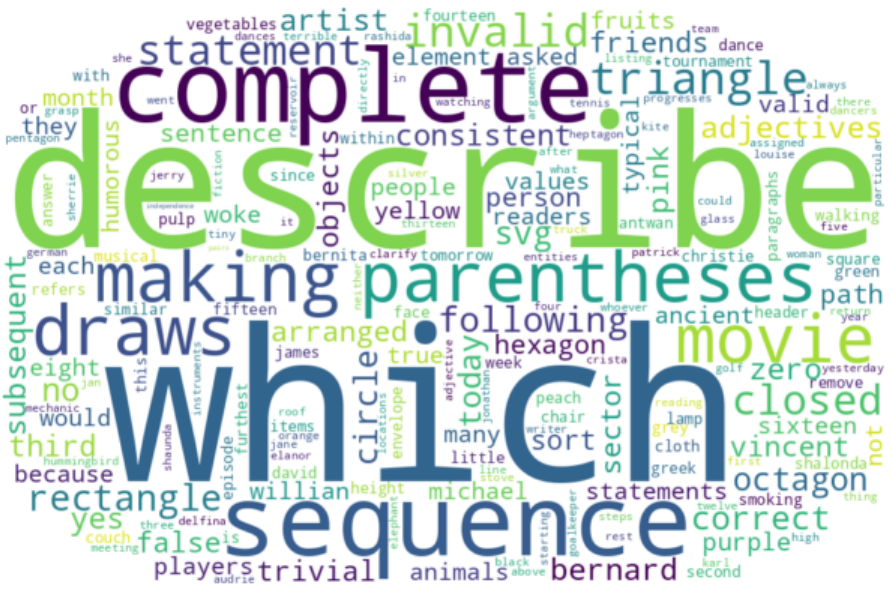}
        \caption{BBH}
        \label{fig:sub2}
    \end{subfigure}\hfill
    \begin{subfigure}{0.33\textwidth}
        \centering
        \includegraphics[width=\linewidth]{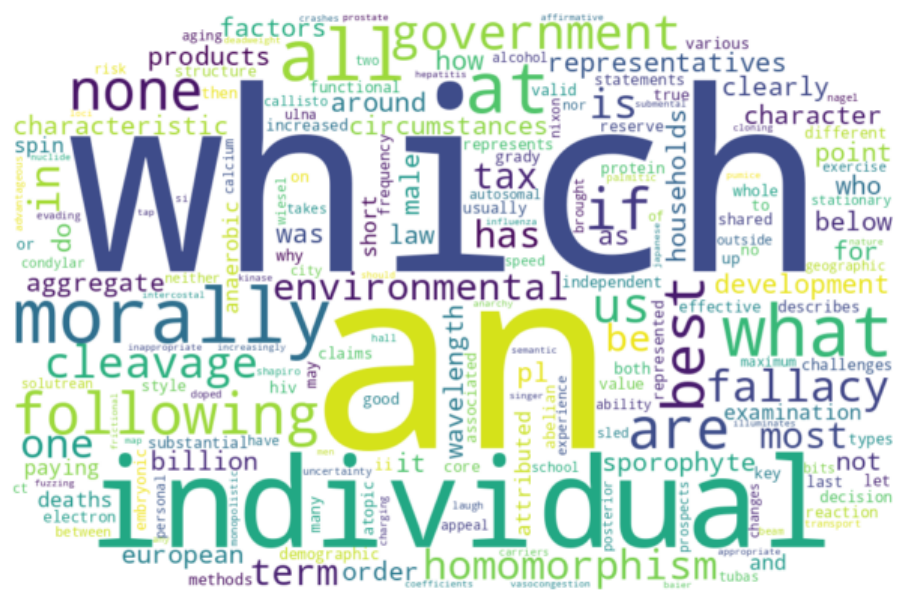}
        \caption{MMLU}
        \label{fig:sub3}
    \end{subfigure}
    \caption{Statistic of words edited in \texttt{R$^2$ATA}.}
    \label{fig:words statistic}
\end{figure*}

\begin{figure*}[ht]
    \centering
    \begin{minipage}{0.58\textwidth}
    \begin{subfigure}{0.48\textwidth}
        \centering
        \includegraphics[width=\linewidth]{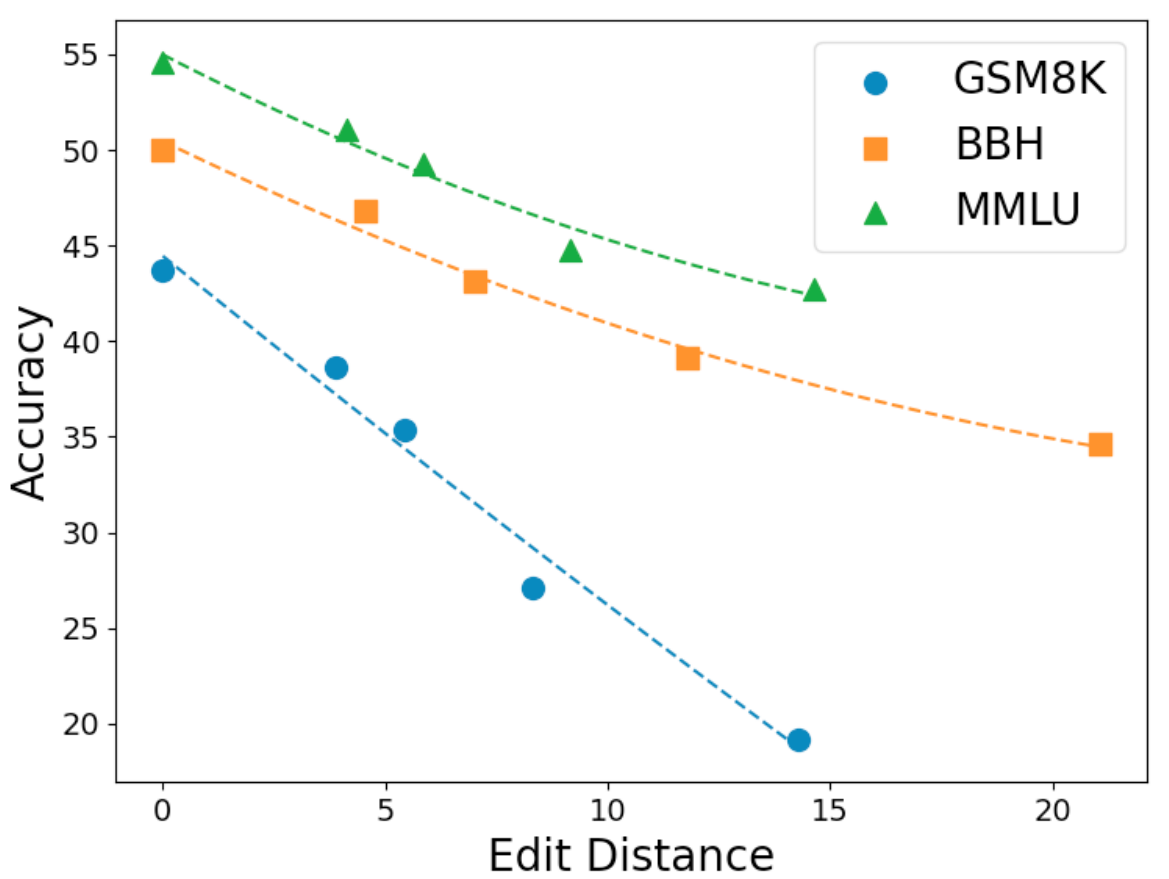}
        \caption{Edit Distance. From left to right, each data point represents 0, 1, 2, 4, 8 edits respectively.}
        \label{fig:edit}
    \end{subfigure} \hfill
    \begin{subfigure}{0.48\textwidth}
        \centering
        \includegraphics[width=\linewidth]{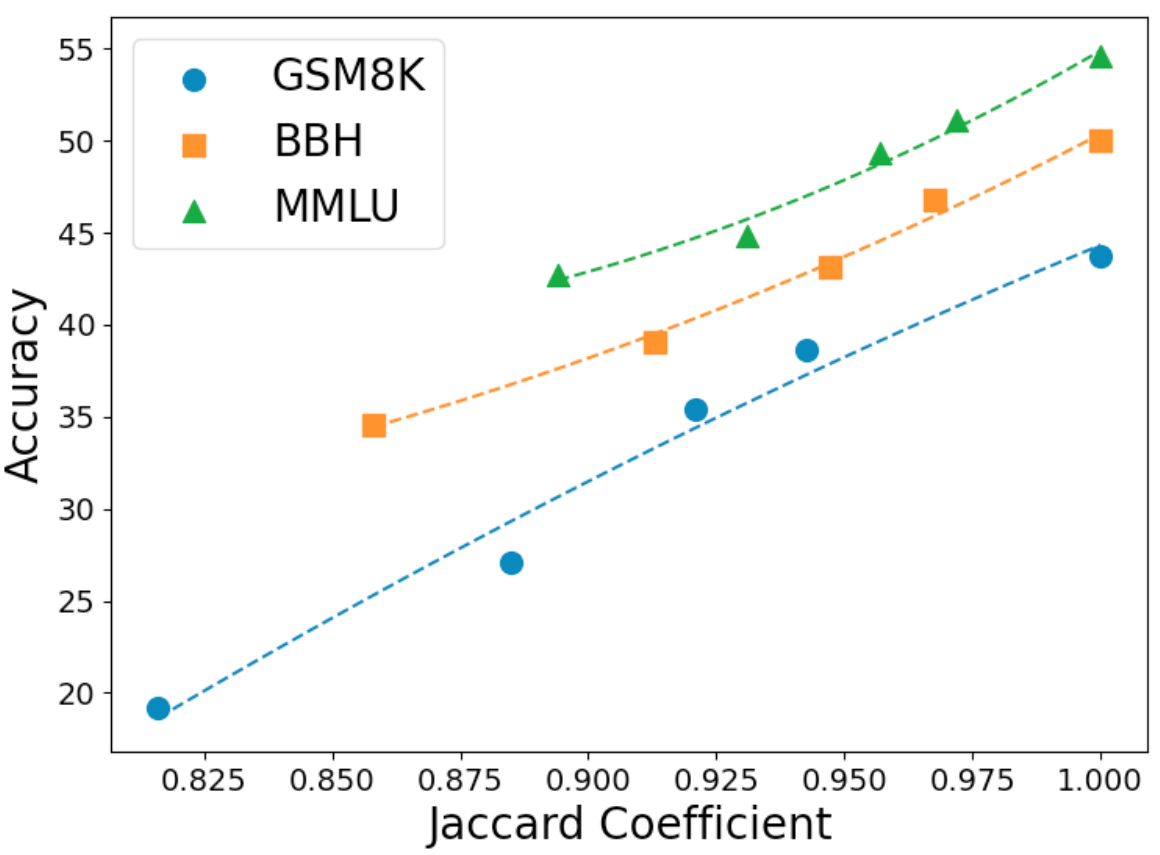}
        \caption{Jaccard Coefficient. From left to right, each data point represents 8, 4, 2, 1, 0 edits respectively.}
        \label{fig:jac}
    \end{subfigure}
    \caption{Effects of adversarial edits at the token level.}
    \label{fig:token}
 \end{minipage}\hfill \hfill
 \begin{minipage}{0.4\textwidth}
    \centering
    \includegraphics[width=\linewidth]{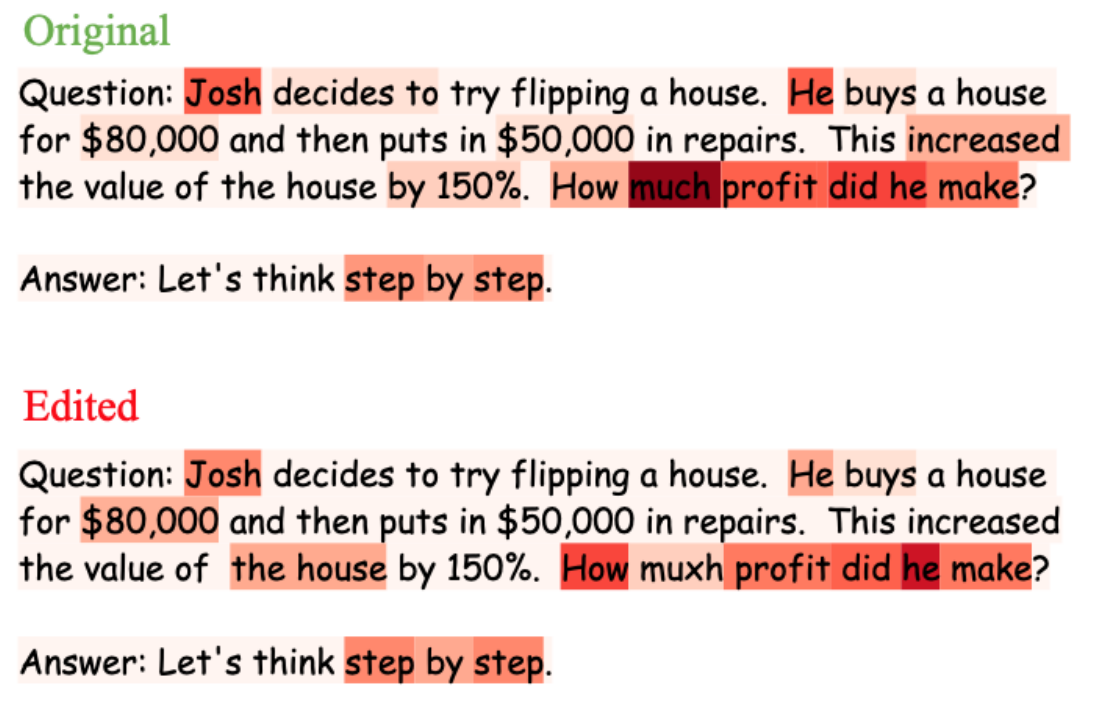} 
    \caption{Visualizing attention changes before and after adversarial attacks. }
    \label{fig:attention}
\end{minipage}
\end{figure*}

\subsection{R$^2$ATA Analysis}

The \texttt{R$^2$ATA} benchmark is analyzed at various levels to provide comprehensive insights into the types and patterns of typographical errors that impact model performance.

\paragraph{Type of Edited Words.}
Figure \ref{fig:wordtype_distribution} illustrates the distribution of edited word types across all three datasets. The data reveals that nouns are the most frequently edited word type, accounting for 48.9\% of the edits. Verbs follow at 16.7\%, and adjectives at 14.9\%. This distribution reflects the significant roles these word types play in conveying meaning. Nouns, as primary subjects and objects, are often targeted for edits due to their substantial semantic weight, which can profoundly alter sentence meaning and context. Verbs, crucial for actions and states, similarly impact sentence meaning when modified. Adjectives, providing descriptive nuances, can subtly change the tone or implication of text upon editing. In contrast, stop words such as conjunctions and prepositions primarily contribute to grammatical structure rather than semantic content, making them less frequently edited and thus less impactful on overall meaning. This goes to show that models need to be more robust to subject perturbations to ensure more robustness to these typographical errors.

\paragraph{Edited Words Statistics.}

Figure \ref{fig:words statistic} shows the word cloud of edited words with size reflecting edit frequency. To ensure a fair comparison, we applied Inverse Document Frequency (IDF) normalization, calculated using:
$\text{IDF}(t) = \log \left( \frac{N}{df_t} \right)$,
where \( t \) is the term, \( N \) is the total number of prompts, and \( df_t \) is the number of prompts containing the term \( t \). 

We adjust each word's frequency by multiplying it with its IDF weight to highlight words disproportionately edited relative to their overall frequency. 
In the GSM8K dataset, frequent edits of words like ``many,'' ``people,'' ``much,'' ``two,'' ``each,'' and ``total'' suggest their semantic importance in mathematical problems due to their inherent complexity and the model's sensitivity to linguistic patterns and numerical expressions. Figures \ref{fig:words statistic}(b) and \ref{fig:words statistic}(c) show word clouds from BBH and MMLU datasets, highlighting words like ``describe,'' ``which,'' ``complete'' for BBH, and ``individual,'' ``an,'' ``which,'' ``all,'' and ``morally'' for MMLU, which cover diverse topics compared to GSM8K's focus on math. The minimal presence of stop words among frequently edited words indicates that edits target content-bearing words, suggesting that \texttt{ATA} edits aim to disrupt the text's logical flow, coherence, or semantics, thus strategically influencing the model's reasoning abilities.

\vspace{-0.2cm}

\paragraph{Impact on the Token Level.}
Figure \ref{fig:edit} illustrates how accuracy varies with edit distance for adversarially edited prompts across three datasets: GSM8K, BBH, and MMLU. Meanwhile, Figure \ref{fig:jac} shows how accuracy varies with the Jaccard coefficient, with each data point representing 0, 1, 2, 4, and 8 edits. It is evident that even a small number of edits leads to a substantial increase in edit distance, resulting in a significant decline in accuracy. However, despite this increase in edit distance, the Jaccard coefficient remains relatively stable, consistently exceeding 0.8 across all edits. This high degree of similarity between the edited and original prompts suggests that the edits are likely imperceptible to humans, underscoring the challenge of detecting adversarial modifications. 

\vspace{-0.2cm}

\paragraph{Impact on Attention}
Figure \ref{fig:attention} illustrates the changes in attention distribution before and after an adversarial attack on a question. In the original question, attention was focused on critical words such as ``much,'' ``increased,'' and ``by 150\%''. However, after the question was edited, there was a noticeable shift in attention. For instance, the attention on ``much'' decreased significantly due to it being altered to ``muxh''. Similarly, attention on ``increased'' and ``by 150\%'' was entirely lost. Instead, the attention was redirected to irrelevant words like ``the house''. This misallocation of attention led to errors in the reasoning steps, as the model focused on less important parts of the text, thereby compromising its ability to understand and answer the question correctly. The detailed implementation code for attention calculation using PyTorch is shown in Appendix \ref{sec:appen_code}.

\section{Related Work}



Textual Adversarial Attacks have garnered significant attention due to their ability to exploit vulnerabilities in LLMs. These attacks, which manipulate input text to mislead models into incorrect predictions or misleading responses, have been studied extensively at various levels of input granularity: character-level \citep{gao2018black, Li_2019, pruthi2019combating}, word-level \citep{ garg2020bae, jin2020bert,zhou2024mathattack}, sentence-level \citep{shi2023large, xu2024an, turpin2024language, lanham2023measuring} and semantic-level \citet{zhu2023promptbench, parcalabescu2023measuring}, as noted by \citet{zhu2023promptbench}. However, these approaches often generate adversarial examples that are easily detectable by human users, limiting their real world applicability. Our approach instead introduces imperceptible modifications to prompts similar to \citet{brown2018unrestrictedadversarialexamples, Richards2021}, offering a more realistic assessment of adversarial risks. 

Furthermore, while some defenses address related threats, such as malware detection adversaries~\citep{fleshman2018non, Romeo2018}, they operate in more constrained spaces and do not directly apply to the nuanced edits~\citep{lowd2005good} that we explore.

\section{Conclusion}

This study examined the robustness of LLMs to typographical errors using the \texttt{ATA} algorithm and the \texttt{R$^2$ATA} benchmark. By focusing on imperceptible, real-world attacks in NLP, our work fills a key gap in adversarial research, moving beyond the artificial constrains of prior approaches and offering insights into more practical vulnerabilities in LLMs. Our findings show that even minor typographical changes significantly reduce model accuracy. Specifically, we observe that adversarial prompts from Mistral-7B similarly affect larger models like Vicuna-13B, Vicuna-33B, and Mixtral-8$\times$7B, indicating that both smaller and larger models are vulnerable. This highlights the need for improved robustness in LLMs against typographical errors. The \texttt{R$^2$ATA} benchmark is a valuable tool for developing more resilient models capable of reliable performance despite minor errors, emphasizing the critical need for robust defense mechanisms in future LLMs.

\section*{Limitation}

Our algorithm primarily focuses on typographical errors common in languages that use alphabets and whitespaces, such as English. This excludes languages with different writing systems, such as Chinese, where typographical errors may involve character substitutions or stroke omissions. The typographical errors considered may not cover all possible real-world scenarios. For instance, whitespace errors only apply to languages that use spaces, while letter addition and deletion errors are relevant only to alphabetic languages. Therefore, future research should extend the scope to encompass a broader range of linguistic diversity to ensure the applicability of findings across various languages and writing systems. Exploring language-specific modifications will provide a more comprehensive understanding of LLM robustness across diverse linguistic contexts. Developing and testing adversarial attacks tailored to these languages will help in creating more universally resilient language models. Additionally, our evaluation primarily relies on open-source and commercially available LLMs due to accessibility constraints. While the \texttt{R$^2$ATA} benchmark effectively demonstrates vulnerabilities in these models, the performance of many closed-source LLMs remains unexplored.

\section*{Acknowledgements}

This research is partially supported by the National Research Foundation Singapore under the AI Singapore Programme (AISG Award No: AISG2-TC-2023-010-SGIL) and the Singapore Ministry of Education Academic Research Fund Tier 1 (Award No: T1 251RES2207). Additionally, the computational work for this article was partially performed on resources of the National Supercomputing Centre, Singapore.

\bibliography{main}
\bibliographystyle{acl}

\appendix

\section{Appendix}
\subsection{Examples of Edits by \texttt{ATA}}\label{sec:appen_examples}

\paragraph{GSM8K } Table \ref{appendix:GSM8K} shows an example of typographical errors generated on a GSM8K question.

\begin{table}[ht]
  \centering
  \footnotesize
  \scalebox{0.85}{
    \begin{tabular}{lp{6cm}}
      \toprule
      \textbf{Original} & \textbf{Question: Archie buys beef jerky that comes 30 sticks to a bag and costs \$18.00 per bag.  If Archie buys 1 bag while they're \$3.00 off a bag, how much will each stick of jerky cost in cents?\textbackslash nAnswer: Let's think step by step.} \\\midrule
      ATA-1 & Question: Archie buys beef jerky that comes 30 sticks to a bag and  costs \$18.00 per bag.  If Archie buys 1 bag while they're \$3.00 off a bag, how much will each stick of jerky cost in cents?\textbackslash nAnswer: Let's think step by step.\\\midrule
      ATA-2 & Question: Archie buys beec jerky that comes 30 sticks to a bag and  costs \$18.00 per bag.  If Archie buys 1 bag while they're \$3.00 off a bag, how much will each stick of jerky cost in cents?\textbackslash nAnswer: Let's think step by step. \\\midrule
      ATA-4 & Question: Archie buys beec jerky that comes 30 sticks to a bag and  costs \$18.00 per bag.  If Archie buys 1 bag while they're \$3.00 off a bag, how much will each stick of jer kg  cost in cents?\textbackslash nAnswer: Let's think step by step. \\\midrule
      ATA-8 & Question: Archie buys beec jerky that comes 30 sticks to a bag and  costs \$18.00 per bag.  If  Archie buys 1 bag while they're \$3.00 off a bag, how much will eacn stick of  jer  kg  cost in cents?\textbackslash nAnswer: Let's think step by step. \\
      \bottomrule
    \end{tabular}
  }
  \caption{An example of typographical errors generated on a GSM8K question.}  
  \label{appendix:GSM8K}
\end{table}

\paragraph{BBH} Table \ref{appendix:BBH} shows an example of typographical errors generated on a BBH question.

\begin{table}[ht]
  \centering
  \footnotesize
  \scalebox{0.85}{
    \begin{tabular}{lp{6cm}}
      \toprule
      \textbf{Original} & \textbf{Q: Is the following sentence plausible? "Petr Cech was safe at first."\textbackslash nA: Let's think step by step.} \\\midrule
      ATA-1 & Q: Is the following sentence plausible? "Petr Cech was szfe at first."\textbackslash nA: Let's think step by step.\\\midrule
      ATA-2 & Q: Is the following sentence plausible? "Petr Cech was szfe at  first."\textbackslash nA: Let's think step by step.\\\midrule
      ATA-4 & Q: Is the folllwing sentence plausible? "Petr Cech was  szfe at  first."\textbackslash nA: Let's think step by step. \\\midrule
      ATA-8 & Q: Is the follwing sntence plausible? "Petr Cech was  szfe at  firsst."\textbackslash nA: Let's think step by step. \\
      \bottomrule
    \end{tabular}
  }
  \caption{An example of typographical errors generated on a BBH question.}  
  \label{appendix:BBH}
\end{table}

\paragraph{MMLU} Table \ref{appendix:MMLU} shows an example of typographical errors generated on a MMLU question.

\begin{table}[ht]
  \centering
  \footnotesize
  \scalebox{0.85}{
    \begin{tabular}{lp{6cm}}
      \toprule
      \textbf{Original} & \textbf{Q: Which of these should an objective NOT be?\textbackslash n(A) Broad (B) Achievable (C) Measurable (D) Time-bound\textbackslash nA: Let's think step by step.} \\\midrule
      ATA-1 & Q: Which of these should an objective NOT be?\textbackslash n(A) Broad (B) Achievable (C) Measurable (D) Tie-bound\textbackslash nA: Let's think step by step. \\\midrule
      ATA-2 & Q: Which of these should an objective NOT be?\textbackslash n(A) Brod (B) Achievable (C) Measurable (D) Tie-bound\textbackslash nA: Let's think step by step. \\\midrule
      ATA-4 & Q: Which of these should an object ve NOT be?\textbackslash n(A) Brod (B) Achievable (C) Measurable (D) Tie-bound\textbackslash nA: Let's think step by step. \\\midrule
      ATA-8 & Q: Which of thee shoulld an object ve NOT be?\textbackslash n(A) Brod (B) Achievable (C) Me aaurable (D) Tie-bound\textbackslash nA: Let's think step by step. \\
      \bottomrule
    \end{tabular}
  }
  \caption{An example of typographical errors generated on a MMLU question.}  
  \label{appendix:MMLU}
\end{table}


\subsection{Calculation of Attention Weights}\label{sec:appen_code}

We obtained the attention weights using the Huggingface library. We obtain from specifically the last attention layer. Because there are 16 attention heads, we chose to perform mean pooling on the attention weight matrix and obtained the attention of all the words with respect to the last token in the user input.

\begin{figure}[ht]
\vspace{-0.2cm}
  \centering
    \includegraphics[width=0.46\textwidth]{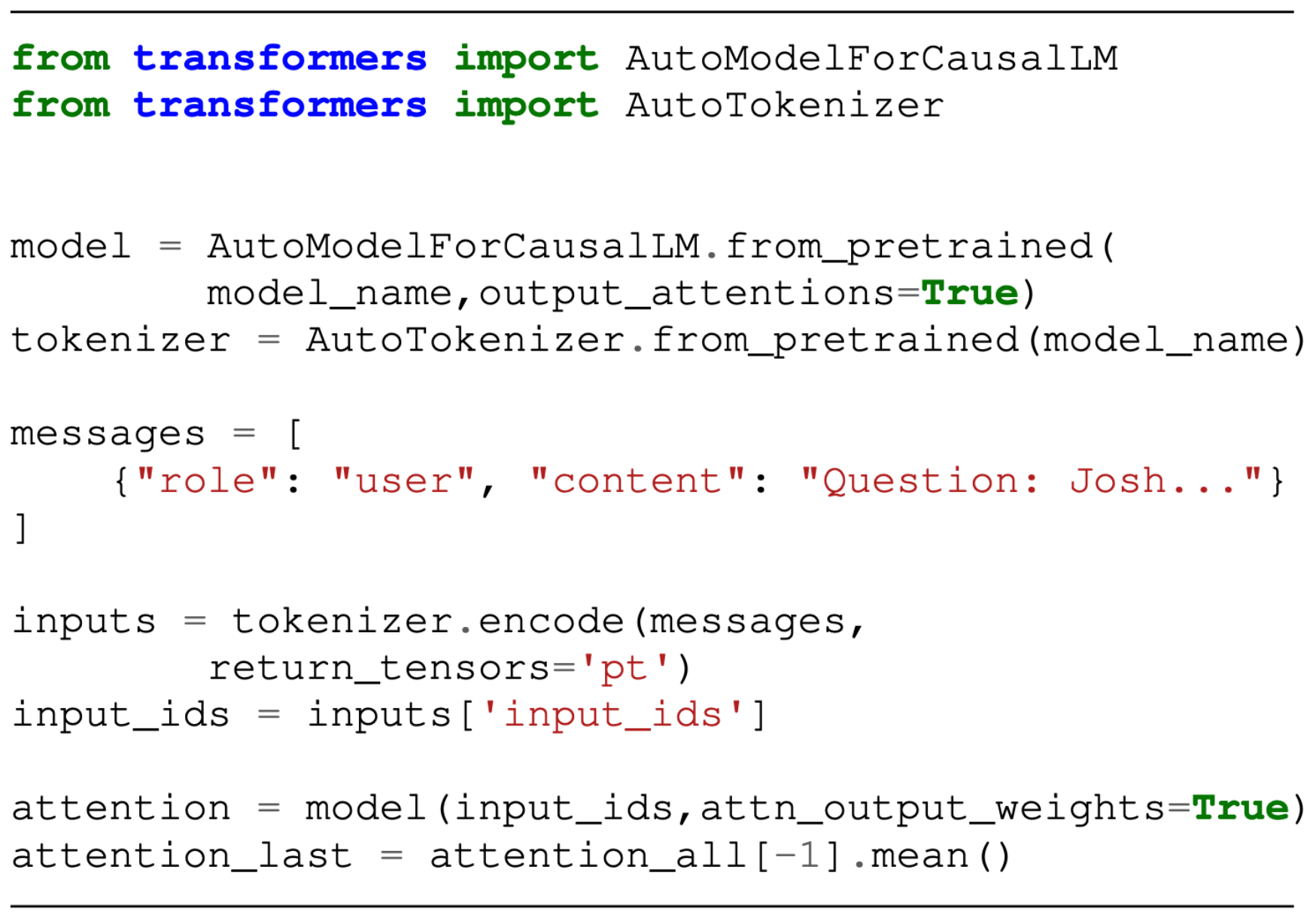}
\vspace{-0.3cm}
\end{figure}


        



\end{document}